\documentclass[runningheads]{llncs}
\usepackage{booktabs}
\usepackage{amsmath}
\usepackage{amsfonts}
\usepackage{graphicx}
\usepackage{placeins}
\usepackage{url}
\usepackage{xcolor}
\usepackage{wrapfig}
\newlength{\originalintextsep}
\usepackage[T1]{fontenc}
\usepackage{graphicx}
\begin{document}
\title{FLEXtime: Filterbank learning to explain time series}
%
%
\author{Thea Brüsch\inst{1,3}\and
Kristoffer Knutsen Wickstrøm\inst{2} \and
Mikkel N. Schmidt \inst{1,3}\and Robert Jenssen \inst{2,3,4} \and Tommy Sonne Alstrøm \inst{1,3}}
\authorrunning{T. Brüsch et al.}
%
\institute{DTU Compute, Technical University of Denmark \and
Department of Physics and Technology, UiT The Arctic University of Norway \and
Pioneer Centre for AI, University of Copenhagen, Denmark \and Norwegian Computing Center, Oslo, Norway}
\maketitle              
\begin{abstract}
State-of-the-art methods for explaining predictions from time series involve learning an instance-wise saliency mask for each time step; however, many types of time series are difficult to interpret in the time domain, due to the inherently complex nature of the data. Instead, we propose to view time series explainability as saliency maps over interpretable parts, leaning on established signal processing methodology on signal decomposition. Specifically, we propose a new method called FLEXtime that uses a bank of bandpass filters to split the time series into frequency bands. Then, we learn the combination of these bands that optimally explains the model's prediction. 
Our extensive evaluation shows that, on average, FLEXtime outperforms state-of-the-art explainability methods across a range of datasets. FLEXtime fills an important gap in the current time series explainability methodology and is a valuable tool for a wide range of time series such as EEG and audio. Code is available at \url{https://github.com/theabrusch/FLEXtime}.

\end{abstract}
\section{Introduction}
Explainability of black-box models is paramount for safe decision-making in critical domains such as health care \cite{amann2020a,beger2024a} and finance \cite{weber2024a}. Although recent years have seen an abundance of explainability methods developed for images \cite{achtibat2023a,fong2019a,Kolek2022,rise,relax}, time series have been overlooked to a higher degree \cite{di2023a}. A possible reason is the inherently complex nature of time series \cite{rojat2021a} which makes it difficult for humans to disentangle salient information. 

\begin{figure}
    \centering
    \includegraphics{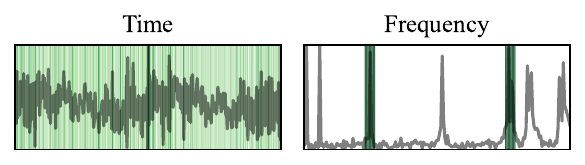}
    \caption{Explainability methods that provide explanations (green heatmaps) in the time domain fail to parsimoniously explain time series data if the salient information is localized in the frequency domain.}
    \label{fig:time_freq}
\end{figure}

Explainability methods, in general, can be divided into local and global explanations. Our work focuses on local methods. Local explainability methods provide explanations for single samples of data. Local methods often produce explanations as saliency maps over input features \cite{lrp,guided-backprop,integratedgradients}, thereby leaving the interpretation of the explanation to the user. Moreover, many methods are based on smoothness and sparsity constraints, assuming localized and sparse information in the explained domain~ \cite{dynamask,fong2019a,Kolek2022}. This behavior may not be suitable for explaining time series, where salient information may often be found in latent feature domains such as the frequency domain \cite{schroeder2023a}. 
Consider, as an example, the case where a class is characterized by the presence of two specific frequency components. This characteristic would be neither localized nor sparse if represented in the time domain. The concept is illustrated in Figure \ref{fig:time_freq} where the explanations in time and frequency domains, respectively, are plotted as heatmaps.

State-of-the-art methods for explaining time series often rely on learnable masks to create attribution maps in the time domain \cite{dynamask,extrmask,timexplusplus,contralsp,timex}. The masks are learned through gradient descent using an objective function that ensures that the model's output on the masked input aligns with the original output while masking out as much of the input as possible. However, while these methods yield high performance on time series where salient information is localized in the time domain, they are unable to describe salient features in the frequency domain. 

In this work, we offer a novel perspective on time series explainability: Instead of saliency maps over the raw input space, we propose learning saliency maps over interpretable parts. Leveraging established signal processing methodology, we decompose the time series using a structured dictionary of interpretable components which serves as a basis for the explanations. Specifically, we propose to use a bandpass filterbank to learn explanations in the frequency domain, but we note that our methods can be adapted to other structured dictionaries. We combine all elements into a framework we refer to as FLEXtime (Filterbank Learning to EXplain time series). FLEXtime provides users the flexibility to define filter banks according to specific domain requirements, enhancing the interpretability of the learned explanations. We show that FLEXtime outperforms our developed baseline FreqMask (a frequency-oriented method based on Dynamask \cite{dynamask}), which learns explanations directly in the discrete Fourier domain. 
To the best of our knowledge, we are the first to develop a method for learnable explanations for time series in the frequency domain. 
Our contributions are as follows:

\begin{itemize}
    \item[$\bullet$] We propose to reimagine time series explainability as explainability over interpretable parts to learn meaningful saliency maps. 
    \item[$\bullet$] We propose FLEXtime, which leverages our proposed methodology by using a filterbank of bandpass filters to learn explanations in the frequency domain.
    \item[$\bullet$] We present the baseline FreqMask, inspired by Dynamask \cite{dynamask}, that computes saliency maps directly on the frequency axis. 
    \item[$\bullet$] We evaluate FLEXtime and FreqMask across a number of synthetic and real datasets with a range of descriptive metrics. FLEXtime outperforms all baselines, achieving the best average rank on faithfulness, robustness, and smoothness. 
\end{itemize}

\section{Previous work on time series explainability}
Explainability for images has been a rapidly evolving field over the past years, whereas explainability for time series remains largely underexplored. While most natural images contain semantics that are easy for humans to decipher, the unintuitive nature of time series data makes interpretation difficult, even for experts \cite{schlegel2019towards}. Thus, time series represent unique challenges for the explainability community. Here, we present an overview of relevant work within the image domain and proceed to give an overview of the time series explainability field.  Generally, explainability methods can be divided into intrinsic vs. post-hoc based on whether the explainability is build into the model, or if the explanation is computed after training. Additionally, explainability methods can be said to be either model-specific or model-agnostic. Model-agnostic methods can be applied to any model, while model-specific methods can only be applied to some models.
We focus on local post-hoc explainability methods. 

\textbf{Explainability for images}
Generally, a number of gradient-based methods have been widely used for image explainability. This category includes Guided backpropagation \cite{guided-backprop}, Layer-wise Relevance Propagation \cite{lrp}, and Integrated Gradients \cite{integratedgradients}, which all use different techniques to distribute the gradient with respect to the model output in the input space. Since these methods require access to model gradients, all of them can be said to belong to the model-specific category. Gradient-based methods have been criticized for their unreliability in robustly determining importance \cite{impossibility}.\\ 
Occlusion-based methods remove part of the input and assess the change in the model output. RISE is a model-agnostic method that samples masks and uses a weighted average to compute saliency maps \cite{rise}. Fong et al. \cite{fong2019a} learn the mask using an optimization criterion. Kolek et al. \cite{Kolek2022} propose a rate-distortion framework to learn the mask. Both of the latter can be said to be model-specific, since they require us to backpropagate gradients from the output through the model to the input. Occlusion-based methods directly observe correlations between input perturbations and model responses.

\textbf{Explainability in the time domain}
Some recent work has focused on developing methods tailored for time series. Model-agnostic approaches include adopting methods such as LIME \cite{lime}, SHAP \cite{shap}, and RISE \cite{rise} to the time series domain \cite{mercier2022a,windowshap,limesegment}. However, the most successful approaches rely on the occlusion-based approach with learnable masks in the input domain. We base our work on these recent advances of learnable masks, noticing that none of these have explored learnable masks in the frequency domain. 

The first method developed specifically for time series, Dynamask \cite{dynamask}, poses a learning objective inspired by Fong. et al \cite{fong2019a} to learn extremal masks directly in the time domain. They adopt the method to the time domain, by adding a smoothness regularizer to avoid sudden jumps in saliency over time, and through dynamic perturbations of the input. 
ExtrMask \cite{extrmask} extends Dynamask by learning the perturbations through a neural network, while ContraLSP additionally applies contrastive learning when learning the perturbations \cite{contralsp}. TimeX recognizes that masking may lead to out-of-domain samples and instead trains a surrogate model that is more robust to masked inputs \cite{timex}. TimeX++ \cite{timexplusplus} uses an information bottleneck to learn the explanations and a neural network to produce in-domain masked samples. 

While ExtrMask, TimeX, and TimeX++ all outperform Dynamask, the addition of neural networks for either mask or perturbation generation adds complexity and need for tuning of several hyperparameters - something that is not trivial to do for explainability tasks. Additionally, none of the methods can produce explanations in other domains than the time domain. 


\textbf{Modeling time series in the frequency domain}
Recently, frequency modeling of time series has received more attention in the deep learning community. First, we examine recent uses in the general deep learning community. Second, we look at the few works that directly target time series explainability in the frequency domain. 

Liu et al. \cite{frequency-aware-masked-autoencoders} train a foundation model using a frequency-aware transformer architecture and show its superiority over models that do not consider the frequency domain. Crabbé et al. \cite{crabbé2024time} show that the frequency domain better captures the data distribution for diffusion models. 
While none of this is directly related to explainability, it does indicate an increasing tendency to exploit frequency representations of time series in deep learning and related fields. 

Few works have focused on explaining time series models in the frequency domain, but there are some notable recent works. 
Vielhaben et al. \cite{vilrp} use virtual inspection layers to propagate relevance from the time domain into the frequency domain. Gradient-based methods, though, have been outperformed by learnable masks in the time domain \cite{dynamask}. Finally, FreqRISE uses sampling of masks to estimate relevance in the frequency domain \cite{freqrise}. However, FreqRISE relies on computationally inefficient sampling and applies masks in the frequency domain by directly zeroing out frequency components, which may lead to artifacts (see Section \ref{sec:artifacts}).

In this work, we build on these recent trends in explaining time series in alternative domains by decomposing the signals into interpretable parts. We leverage the powerful concept of learnable masks to do so, and present a promising new direction for time series explainability that we call FLEXtime. FLEXtime is a local model-specific post-hoc method.

\section{Explainability over interpretable parts}
Instead of viewing the explainability of time series as a saliency map over the input, we propose instead the notion of explainability over interpretable parts. This new approach allows us to provide the user with saliency maps over elements of a more inherently interpretable nature for the time series in question. Additionally, due to the recent success of learnable masks, we propose to learn the saliency map over these interpretable parts. 
The section is organized as follows. First, in Section \ref{sec:dictionary}, we introduce the notion of learning a saliency map over interpretable parts. Section \ref{sec:flextime} proceeds to introduce our proposed method, FLEXtime. Finally, Section \ref{sec:freqmask} establishes a simpler baseline, FreqMask inspired by Dynamask \cite{dynamask}.
\subsubsection{Notation}

Let $X \in \mathbb{R}^{N\times V}$ be a uniformly sampled time series consisting of $N$ time steps and $V$ variables, drawn from the distribution $\mathcal{X}$, with the associated label or output $y \in \mathbb{R}^{C}$ with dimension, $C$. 
Here, $y$ could be any regression or classification variable that is associated with $X$ for the given task. For ease of notation, we will assume $V=1$ without loss of generality, but the approach applies for any $V\in\mathbb{N}$. 

We then assume a (black box) model $\mathbf{f}$ that predicts $\hat{y}$ from $X$, $\mathbf{f}: X \rightarrow \hat{y}$. The goal is now to explain the model's prediction in terms of the input $X$. The explanation should identify the most important features of the input $X$ for $\mathbf{f}$ in predicting $\hat{y}$.

\subsection{Learning a mask over interpretable parts} \label{sec:dictionary}
We now lean on the signal processing methodology on sparse signal representations \cite{theodoridis2020a}. Many applications in signal processing rely on finding a sparse signal representation for the purpose of compression \cite{7457891}, denoising \cite{sparsedenoising}, and higher interpretability \cite{kim2010sparse,10095297,tolooshams2022stable}. The sparse representation is a linear decomposition of the signal, $X$, into a suitable dictionary consisting of interpretable elements, $\{\psi_s\}_{n=1}^S$ \cite{theodoridis2020a}:
\begin{equation} \label{eq:dictionary}
    X = \sum_{s=1}^S \theta_s \psi_s.
\end{equation}
where $\theta_s\in\mathbb{C}$. Notice here that the choice of dictionary and thus interpretable parts will be dependent on the signal in question and could be determined by domain experts. 

In order to explain the prediction of our model $\mathbf{f}$ on input $X$ in terms of our dictionary, we seek a mask $M = (m_{s}) \in [0,1]^S$. The value of $m_{s}$ should indicate the saliency of element $\psi_s$, where $m_{s}$ close to one indicates that element $\psi_s$ is salient, while $m_{s}$ close to zero indicates that element $\psi_s$ is not salient. $M$ is used to mask out elements of $X$ via elementwise multiplication:
\begin{equation}
    X^M = \sum_{s=1}^S \left(m_s \cdot \left(\theta_s \psi_s\right) + \left(1-m_s\right)\cdot  p_s\right).
\end{equation}
Here, $p_s$ is a perturbation applied for dictionary component $s$.
We can obtain the resulting output of $\mathbf{f}$:
\begin{equation} \label{eq:yhat}
   \hat{y}^M = \mathbf{f}(X^M).
\end{equation}

The goal is now to learn the mask $M$ that optimally explains the output of the model. 
We will achieve the optimal mask, $M$, through optimization and thus need a set of desiderata to design the optimization objective. 

For an optimal mask $M$ that has identified all salient information, we expect $\hat{y}^M \approx \hat{y}$. As such, an objective should be to minimize the difference between $\hat{y}^M$ and $\hat{y}$. Generally, we can quantify this difference with some kind of distortion term, $D$. The distortion term, $D$, should be chosen according to the task at hand. I.e., for regression tasks the mean squared error would likely be a good choice, while for classification tasks the cross-entropy would be natural. 
However, minimizing $D$ is not sufficient for providing useful explanations. Consider a mask, $M$ where all elements $m_{s}=1, s=1, \dots, S$. For this mask, we get $\hat{y}^M = \hat{y}$, since $X^M=X$ in this case. 

Therefore, it is also necessary to control the sparsity of the mask. Generally, this is done with a regularization metric $R(M)$. Combining the minimization of $D$ with the maximization of the sparsity of $M$ through $R$ allows us to define the optimization objective:
\begin{equation}
\label{eq:loss}
    \underset{M}{\operatorname{min}}\quad D(\hat{y}, \hat{y}^M) + \lambda R(M),
\end{equation}
where $\lambda$ controls the trade-off between $D$ and $R$.

Generally, this objective can be interpreted from a rate-distortion perspective \cite{cover1991a,Kolek2022}. Here, for proper distortion measures, $D(\hat{y}, \hat{y}^M)$ acts as a proxy for the distortion of $X$ caused by $M$, which we want to minimize. Similarly, the rate refers to the amount of signal in $X$ being passed by $M$, which we also want to minimize. 
The objective can be optimized by initializing a mask $M$ and optimizing the values of $M$ via gradient descent \cite{fong2019a}.

\subsection{Filterbank Learning to EXplain time series} \label{sec:flextime}
\begin{figure}[h]
    \centering
    \includegraphics{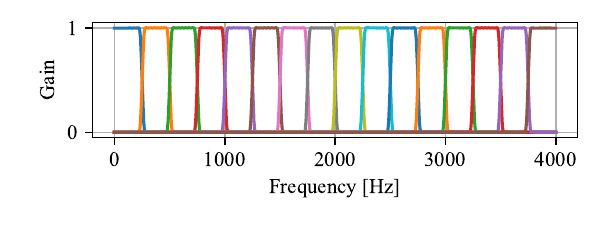}
    \caption{Magnitude response of filterbank with $16$ FIR filters with equal bandwidths.}
    \label{fig:filtbankex}
\end{figure}
How do we decide which dictionary to use to represent $X$? Clearly, this depends on the qualities and application of $X$. However, for many time series applications, the frequency domain is of particular interest \cite{audiomnist,crabbé2024time}. Most natural signals do not carry frequency content on a single frequency but rather across a band of frequencies in an area of interest. We therefore propose composing the dictionary as a filterbank of bandpass filters, each with impulse response $h_l$, to split the signal into appropriate frequency bands:
\begin{equation}
    X \approx \sum_{l=1}^L h_l \ast X,
\end{equation}
where $\ast$ denotes the convolution operation. 
Specifically, in this paper, we propose using a filter bank of $L$ finite impulse response (FIR) filters that split the frequency axis into equally sized bins. This is chosen since FIR filters are stable and can easily be designed to have a linear phase response \cite{oppenheim2010a}. These properties allow for easy automatic design for a variety of datasets. An example of such a filterbank is shown in Figure \ref{fig:filtbankex}.

The filterbank design is inspired by the analysis filterbank framework such as \cite{crochiere1983multirate,vaidyanathan1987a}. Since we apply sparse masks, we do not need perfect reconstruction of our signals. Therefore, we use the window method to design the FIR filters and leave the number of filters, $L$, and filter length, $N_{h}$, as hyperparameters. In practice, however, the choice of filterbank could depend on the domain of choice. For audio, a natural choice could be an octave filterbank, where the frequency axis is divided into progressively wider frequency bands to more closely resemble human acoustics processing \cite{proakis1992a}. 

Independent of the design choice of the filterbank, the mask can now be learned as a combination of bandpass filtered versions of $X$. 
Combining the notation with the notation in Section \ref{sec:dictionary}, we can produce masked versions of $X$ by masking out specific filters:
\begin{equation} \label{eq:filtermask}
    X^M = \sum_{l=1}^L \left(m_{l} \cdot \left(h_{l} \ast X\right) + \left(1-m_{l}\right)\cdot p_{l}  \right).
\end{equation}
Again, $p_{l}$ denotes any applied perturbation. Generally, we set $p_l=0$, but future work might investigate relevant perturbations to include. 

\begin{figure*}[t]
    \centering
    \includegraphics[width=\linewidth]{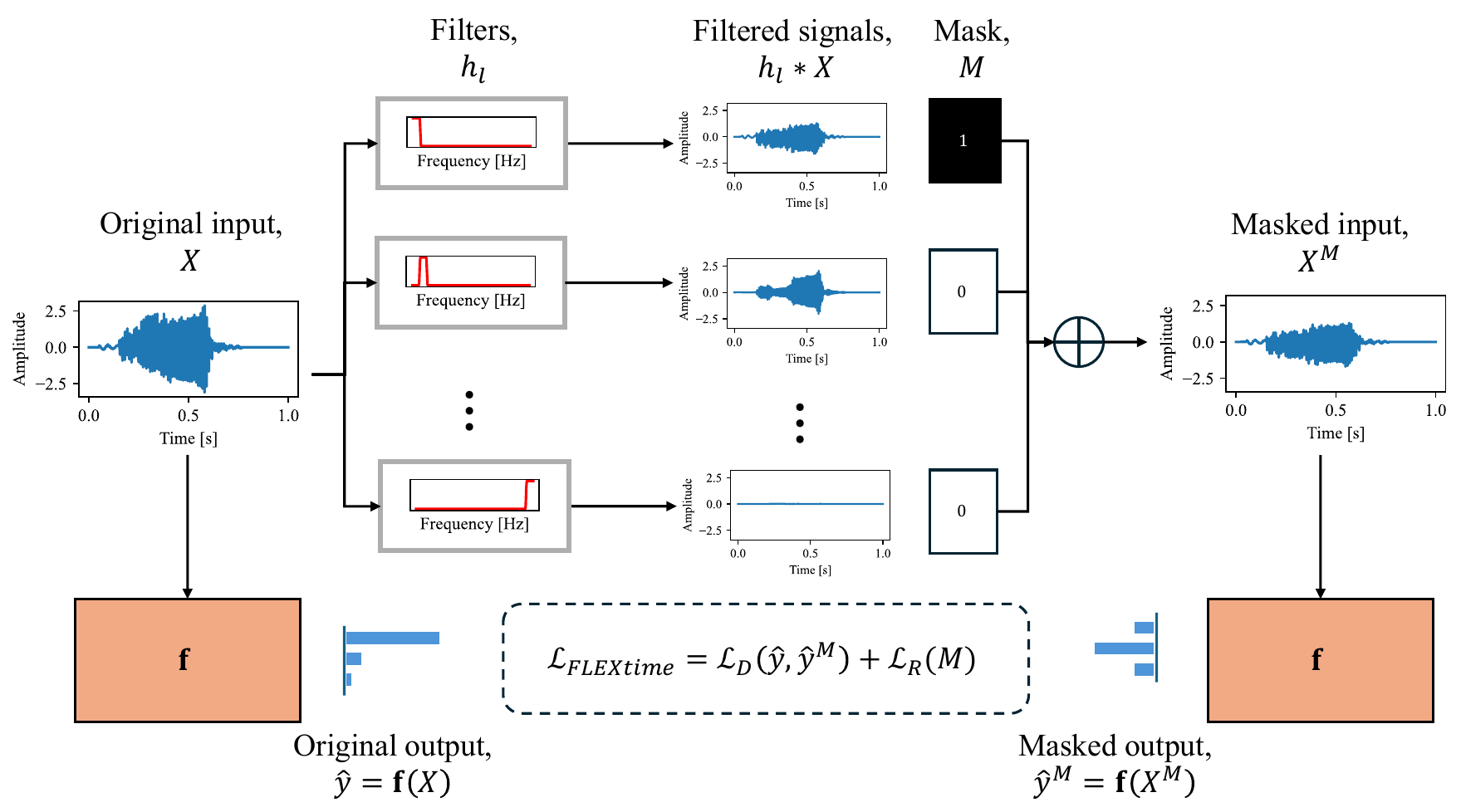}
    \caption{\textbf{FLEXtime}: FLEXtime uses a filterbank to split the signal into frequency bands of a suitable bandwidth. It then optimizes a mask which chooses the frequency bands that best explains the signal $X$ in terms of the prediction $\hat{y}$ of black box $\mathbf{f}$. $\mathbf{f}$ is only used for inference and thus frozen during optimization.}
    \label{fig:flex}
\end{figure*}

We call this method Filterbank Learning to EXplain time series (FLEXtime). The FLEXtime framework is shown in Figure \ref{fig:flex}.



\subsubsection{The FLEXtime learning objective}
In order to learn the mask, $M$, we need to define an appropriate learning objective. This is done by defining appropriate expressions to insert into \eqref{eq:loss}. Since we are focusing on classification tasks, we use the cross-entropy loss to measure the distortion caused by the mask:
\begin{equation} \label{eq:distortion}
    \mathcal{L}_D(\hat{y}, \hat{y}^M) = -\sum_{c=1}^C \log{\left(\hat{y}_c^M\right)}\hat{y}_c,
\end{equation}
where $\hat{y}_c^M$ is obtained using \eqref{eq:filtermask} and \eqref{eq:yhat}.
To ensure that the method only focuses on the class, $l$, we wish to explain, we set all values of $\hat{y}_{c\neq l} = 0$ prior to computing $D(\hat{y}, \hat{y}^M)$.

For the sparsity constraint, we use the $\ell_1$-norm. Additionally, to control the sparsity of the explanation, we introduce a ratio parameter, $r\in[0,1]$, which acts as a threshold below which the sparsity constraint is ignored. Thus, we arrive at the following regularization metric:
\begin{equation}\label{eq:rate}
    \mathcal{L}_R(M) = \max{\left(\frac{||M||_1}{L}-r,0 \right)},
\end{equation}
where $L$ is the number of filters and therefore the length of the mask.
The ratio parameter, $r$, will typically be chosen as a small value in the range of $0.05-0.15$, but can be tuned according to the user's need. We then arrive at the following objective function:
\begin{equation} \label{eq:flextime}
    \mathcal{L}_{FLEXtime} = \mathcal{L}_{D}(\hat{y}, \hat{y}^M) + \lambda \mathcal{L}_R(M),
\end{equation}
where $\lambda$ determines the balance between the two loss terms.

\subsection{Learning the mask in the Fourier domain} \label{sec:freqmask}
In FLEXtime, we use a filterbank of bandpass filters to learn explanations in the frequency domain. A simple baseline to measure this approach against, would be a masking strategy directly on the spectrum of the signal, i.e. removing individual frequency components. This can be achieved using another linear transform, namely the discrete Fourier transform (DFT). 
The DFT uses Fourier analysis to decompose the signal into sinusoids of different frequencies. As such, the Fourier spectrum describes how much energy each frequency component holds in the signal of interest. Using the inverse DFT, we can reconstruct the original signal as:
\begin{equation}
    x_{n}= \frac{1}{N}\sum_{k=0}^{N-1} a_{k} e^{i\omega_{k}}e^{i2\pi\frac{k}{N}n}, 
\end{equation}
where $n$ is the $n$-th time step and $N$ is the total length of the signal. $a_{k}$ and $\omega_{k}$ denote the magnitude and phase, respectively, of the $k$-th frequency component. Thus, we again arrive at a formulation equivalent to \eqref{eq:dictionary}.
We can then create masked versions of $X$ as:
\begin{equation}
\label{eq:fouriermask}
    x^M_{n}= \frac{1}{N}\sum_{k=0}^{N-1} \left(m_{k} \cdot \left(a_{k} e^{i\omega_{k}}e^{i2\pi\frac{k}{N}n}\right) + \left(1-m_{k}\right) \cdot p_{k}\right),
\end{equation}
where $p_k$ denotes a perturbation added to the masked out components in the frequency domain.
For real signals, the DFT is even symmetric, which implies that $a_k=a_{-k \mod N}$, creating an effective mask size of length $N/2+1$ by letting $m_k=m_{-k \mod N}$.


\subsubsection{The FreqMask method}
Inspired by the time series explainability method, Dynamask \cite{dynamask}, we now present FreqMask, masks directly in the frequency domain as described above. 

In accordance to the original Dynamask approach, we employ a windowed fading moving average perturbation.
Naturally, here the windowed moving average is done in the frequency domain. By averaging the amplitudes $a_k$ in a window centered at the current frequency, we obtain:
\begin{equation}
p_k =\frac{1}{2 W+1} \sum_{k^{\prime}=k-W}^{k+W} a_{k^{\prime}},
\end{equation}
where $2W$ is a chosen window size. We set it to the default value of $W=10$. 
For easy comparability to FLEXtime, we use \eqref{eq:distortion} and \eqref{eq:rate} for the distortion and rate terms in the optimization objective. In \eqref{eq:rate}, we replace $L$ with $N/2+1$, the size of the mask.

Finally, inspired by Dynamask \cite{dynamask}, we use a smoothness constraint to avoid sudden jumps in saliency:
\begin{equation} \label{eq:smoothness}
    \mathcal{L}_{S}(M) = \sum_{k=0}^{N-1} \left|m_{k+1}-m_{k}\right|.
\end{equation}
We therefore arrive at the following objective for the FreqMask method:
\begin{equation}
\label{eq:freqmaskloss}
    \mathcal{L}_{FreqMask}= \mathcal{L}_{D}(\hat{y}, \hat{y}^M) + \lambda_R \mathcal{L}_R(M) + \lambda_S \mathcal{L}_{S}(M).
\end{equation}

FreqMask comes with a few significant limitations. By directly altering frequency components, FreqMask can only remove frequency content that aligns with the Fourier transform coefficients, e.g. if we have a signal of a one second duration, this is precisely the integer frequencies. Additionally, zeroing out frequency components can cause artifacts in the signal (see Section \ref{sec:artifacts} for an example). Finally, it is worth noting that the artifacts become more prominent if the DFT does not have the same length as the signal \cite{proakis1992a}. 

\section{Experimental setup}
Here, we describe practical details of the filterbank design and the critical components of our quantitative analysis, such as datasets, baselines, and metrics.
\subsection{Hyperparameters of FLEXtime and FreqMask}
For both FLEXtime and FreqMask, we always set $\lambda_{R}=\lambda_{S}=1$ since we find that this balances the loss terms sufficiently and this choice limits the necessary tuning of hyperparameters. We always optimize FLEXtime and FreqMask using gradient descent for 1000 iterations with a step size of 1. When evaluating FLEXtime, we compute the collective frequency response of the filterbank after applying the learned mask. 

For FLEXtime, we use a filterbank of $L$ FIR filters that splits the frequency axis into equally sized bins. This leaves us with a number of design choices: the number of filters $L$, the length of the filters $N_{h}$, and the threshold $r$ below which we ignore the sparsity constraint. For FreqMask, we need to choose the threshold $r$. 
Ideally, these parameters are chosen by domain experts who have knowledge of which resolution and sparsity may be necessary and sufficient. 
\setlength{\originalintextsep}{\intextsep}
\setlength{\intextsep}{0pt}
\begin{wraptable}{r}{6cm}
    \centering
    \caption{Hyperparameters of FLEXtime and FreqMask for each dataset.}
    \begin{tabular}{l|r|r|r|r}
    \toprule
                & \multicolumn{3}{|c|}{FLEXtime} & \multicolumn{1}{|c}{FreqMask} \\ 
        Dataset & \multicolumn{1}{|c|}{$L$} & \multicolumn{1}{|c|}{$N_{h}$} & \multicolumn{1}{|c|}{$r$} &\multicolumn{1}{|c}{$r$}\\ \hline
        Gender & 128 & 501 & 0.10 & 0.10 \\
        Digit & 128 & 501 & 0.10 & 0.05 \\
        PAM & 32 & 95 & 0.10 & 0.05 \\
        Epilepsy & 32 & 75 & 0.10 & 0.05 \\
        ECG & 64 & 105 & 0.05 & 0.05 \\ 
        SleepEDF &  256 & 901 & 0.10 & 0.10 \\ \bottomrule
    \end{tabular}
    \label{tab:flextime_hyper}
\end{wraptable} 
\setlength{\intextsep}{\originalintextsep}
Here, we instead use a cross-validation scheme to choose the optimal parameters based on faithfulness.Specifically, we loop through each split of each dataset and randomly sample 100 datapoints from each validation split. 
We then do a grid search over the number of filters, $L$, the filter length, $N_{h}$ and the sparsity controlling ratio, $r$ for 
FLEXtime. For FreqMask, we only need to choose the sparsity controlling ratio, $r$. 
We choose the hyperparameters that give the best faithfulness score on a 10\% level. Given a tie, we choose the set with the lowest complexity. The chosen hyperparameters are shown in Table \ref{tab:flextime_hyper}.

\subsection{Datasets}
Here, we present the datasets on which we evaluate the explainability methods.
\subsubsection{Synthetic dataset}
We generate a synthetic dataset with known localized salient information. We divide the frequency axis into $K=32$ equally sized regions. We add frequency content within these regions, modulated by a Voigt profile \cite{Tepper_Garc_a_2006}. This allows us to model a dataset with content that more closely resembles real-life signals, where frequency content is found in bands. We sample $B$ of the $K$ bins, where $B\sim \mathcal{U}(1,10)$. Each bin, $b$, has a start frequency, $f_{b, start}$, and an end frequency, $f_{b, end}$. We linearly distribute $20$ frequencies, $f_b \in [f_{b, start}, f_{b, end}[$ on which we add frequency content. We therefore generate the data within each sampled bin, $b$, as:
\begin{equation}
    x^b_n = \sum_{f_b \in [f_{b, start}, f_{b, end}[} \left(a^b_{f_b} \sin\left(\frac{2\pi n}{N {f_b}} + \psi^{b} \right) \right),
\end{equation}
where $a^b_f$ is determined by the Voigt profile with peak location sampled within the bin. $\psi^b$ is the phase, which is sampled from a uniform distribution for each bin, $\psi^b \sim \mathcal{U}\left(0, 2\pi\right)$. 
Finally, we create $x_n$ by summing the frequency content of all bins:
\begin{equation}
    x_n = \sum_{b=1}^B X^b_n + \epsilon,
\end{equation}
$\epsilon$ is random noise sampled from a normal distribution, $\epsilon\sim\mathcal{N}\left(0, \sigma^2\right)$.
We then choose four regions as salient, where the class labels are the powerset of the salient regions, i.e. there is a total of 16 classes. We generate 5 training sets of $10^4$ samples and train a convolutional neural network on each split. We then generate 5 balanced test sets of $992$ samples (i.e. 62 samples from each class) on which we test the methods.

\subsubsection{Real-life datasets}
We evaluate on five different datasets with a total of six different tasks when testing the model in a real-life setting. 

Adhering to previous research on time series explainability \cite{timexplusplus,timex}, we use \textbf{PAM} for human activity recognition \cite{pam}, \textbf{MIT-BIH (ECG)} for arrythmia detection \cite{mitecg}, and \textbf{Epilepsy (EEG)} for seizure detection \cite{epilepsy} datasets. We use the same transformer-based model architectures and data preprocessing pipelines as in \cite{timex}. All datasets are divided into five splits, and one model is trained for each split. 

Additionally, we use the \textbf{AudioMNIST} dataset \cite{audiomnist}, which contains spoken digits from 0-9. This dataset has been shown to have salient information in the frequency domain \cite{audiomnist,vilrp}. The dataset contains two tasks: gender and digit classification, divided into four and five splits, respectively. We use the same preprocessing pipeline and convolutional model architecture as in the original work \cite{audiomnist}. We train a model for each split. 

Finally, we also include the \textbf{SleepEDFx} dataset \cite{PhysioNet,sleepedf}. The dataset contains whole-night EEG data annotated for sleep stages. Frequency bands are an important discriminator for sleep stages \cite{altalag2019a}.  We divide the subjects into 5 splits to ensure no leakage. We use the same convolutional architecture as for the AudioMNIST dataset to train a sleep staging model for each split. 

For all datasets we sample 1,000 samples from each test set for evaluation of the explainability methods. However, when computing the robustness scores, we use only 100 samples from each test due to the high computational complexity associated with this evaluation. Table \ref{tab:datasets} contains details on the size and dimensions of all datasets as well as the F1 score of the trained models on the sampled test sets. 

\begin{table}[h]
    \centering
    \caption{Overview of the real-life datasets used. The F1 score refers to the performance of the trained models on the dataset in question.}
    \begin{tabular}{l|r|r|r|r|r|r}
    \toprule
    Dataset & $\#$ samples & Length & Dimension & Classes & Fs (Hz) & \multicolumn{1}{|c}{F1 score} \\ \hline
    Gender  & 30,000 & 8,000 & 1 & 2 & 8,000 & $.97(.03)$ \\
    Digit    &30,000 & 8,000 & 1 & 10 & 8,000 & $.96(.01)$ \\
    PAM & 5,333 & 600 & 17 & 8 & 100 & $.88(.02)$ \\
    Epilepsy & 11,500 & 178 & 1 & 2 & 178 & $.95(.01)$ \\
    ECG & 92,511 & 360 & 1 & 2 & 360 & $.93(.05)$ \\ 
    SleepEDF & 92,511 & 3,000 & 1 & 5 & 100 & $.90(.05)$ \\ \bottomrule
    \end{tabular}
    \label{tab:datasets}
\end{table}

\subsection{Baselines}
We use six different XAI baselines to compare our methods. For gradient-based explanations, we use \textbf{saliency} \cite{saliency}, \textbf{gradient times input} (G$\times$I) \cite{gxi}, \textbf{guided backpropagation} (GB) \cite{guided-backprop}, and \textbf{integrated gradients} (IG) \cite{integratedgradients}. We equip all of these with a virtual inspection layer to propagate the explanations into the frequency domain, see \cite{vilrp}. All methods that have been adapted to the frequency domain is marked by an *. Additionally, we compare to \textbf{FreqRISE} \cite{freqrise}, which is directly designed to provide explanations in the frequency domain. Finally, when computing the faithfulness scores (see Section \ref{sec:metrics}), we add a \textbf{random baseline} by randomly sampling frequency components to zero out.

\subsection{Metrics} \label{sec:metrics}
Quantitative evaluation of explainability is challenging due to the lack of ground truth explanations \cite{hedstrom2023metaquantus}, and is an active field of research. Since ground truth explanations are not available, the quality of an explanation can be estimated by measuring different desirable properties \cite{hedstrom2023quantus}. In the time series explainability literature, most evaluations have been limited to only considering few desirable properties \cite{dynamask,timexplusplus,timex}. In this work, we strive towards evaluating a more comprehensive set of desirable properties. These properties that make up the basis for our evaluation metrics are described below.
\begin{figure}[h]
    \centering
    \includegraphics{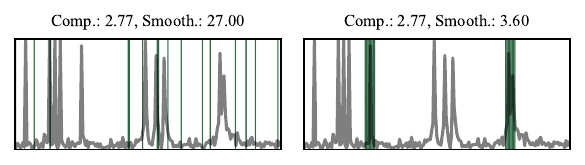}
    \caption{Complexity and smoothness of two different saliency maps.}
    \label{fig:compsmooth}
\end{figure}

\textbf{Localization} Localization measures to what degree the explanation is centered around a known region of interest \cite{loc2,loc1}. For the synthetic dataset with known ground truth, we can directly assess the localization abilities of each method. Here, we follow previous work and compute the area under the recall curve (AUR) and precision curve (AUP), as well as the area under the precision-recall curve (AUPRC) \cite{timexplusplus,timex}.

\textbf{Faithfulness} Faithfulness measures how aligned an explanation is with the prediction of a model \cite{faithfulness2,bhatt2020a}. Specifically, given an explanation, we keep only the 10\% most important features as identified by the explanation. We then measure the mean true class probability of the model on this importance-masked data. For an explanation, where the most important information has been identified, we expect a high mean true class probability. 

\textbf{Complexity} An explanation should be as simple as possible such that it is easy for humans to understand. This can be measured by estimating the complexity of the saliency map \cite{bhatt2020a,complexity2}. We measure the complexity as the entropy of the fractional importance of feature $m_i$ to the overall magnitude following \cite{bhatt2020a}. Low complexity is better, but looking at complexity alone can be misleading. A saliency map that only highlights a single frequency component will have minimal complexity, but might not be informative. Therefore, complexity should be considered in conjunction with metrics that are connected to known regions of interest (localization) or the predictions of the model (faithfulness). 

\textbf{Smoothness} While complexity measures the sparsity of the signal, it yields no information on the smoothness of the saliency map. As such, a saliency map with 10\% of the features marked as salient will have the same complexity independent of whether these features are scattered across the signal or placed in two localized peaks. We therefore suggest considering the smoothness of the saliency maps in conjunction with the complexity. Specifically, we propose using the total variation of the saliency map as a measure of smoothness:
\begin{equation}
    S(M) = \sum_{i = 1}^{N-1}|m_{i+1}-m_{i}|,
\end{equation}
where $M$ is the saliency map with length $N$. We illustrate the difference between complexity and smoothness in Figure \ref{fig:compsmooth} where both saliency maps have the same complexity but vastly different smoothness scores.

\textbf{Robustness} A good explainability method should be robust to small changes in the input \cite{agarwal2022rethinking,faithfulness2}. Here, we measure the robustness of the saliency map using the relative output stability (ROS) \cite{agarwal2022rethinking}. The ROS considers the behavior of the underlying model, by measuring the normalized change in the saliency map in response to small perturbations, \textit{relative} to the change in the model output. We create 10 perturbed inputs by adding normally distributed noise with standard deviation $\sigma=0.05\cdot\sigma_{data}$, where $\sigma_{data}$ is the standard deviation across the test set. The noise is added in the time domain, while the saliency maps are still computed in the frequency domain. We follow the original work \cite{agarwal2022rethinking} and report the log ROS.

\section{Results}

All results are computed across all splits of the data and models and reported as the mean and standard error of the mean across splits. 
We bold and underline the best result and any others whose 95\% confidence interval overlaps with that of the best result. If the second-best result's confidence interval does not overlap with the best result's, we underline it. We compute the rank by ranking all methods by mean within each dataset and computing the average across all.

\subsection{Synthetic data}
\setlength{\originalintextsep}{\intextsep}
\setlength{\intextsep}{0pt}
\begin{wraptable}{r}{6cm}
        \centering
    \caption{Localization$(\uparrow)$ on synthetic data}
    \begin{tabular}{l|r|r|r}
    \toprule
         & AUPRC & AUP & AUR \\ \hline
         Saliency* & .19(.04) & .43(.12) & .08(.01)\\
         G$\times$I*  & .21(.04) & .43(.12) & .07(.01) \\
         GB* & .19(.04) & .42(.12) & .09(.01)\\
         FreqRISE & \textbf{.94}(.02) & .62(.02) &$\mathbf{.81}$(.04)\\
         IG*  & $.62$(.04) & $\mathbf{.99}$(.01) & .09(.01)\\
         FreqMask & .10(.01) & .73(.11) & .22(.09)\\ \hline
         FLEXtime  & $\mathbf{.90}(.09)$ & \underline{.86}(.06) & $\mathbf{.84}$(.07) \\ \bottomrule
    \end{tabular}
    
    \label{tab:synth_voigt_noisy}
\end{wraptable}

The localization scores on the synthetic data are reported in Table \ref{tab:synth_voigt_noisy}. 
We see that the FLEXtime method achieves the highest recall, closely followed by Freq\-RISE, while IG gets a very high precision. Freq\-RISE and FLEXtime have the highest AUPRC with no significant difference. 
\setlength{\intextsep}{\originalintextsep}

\begin{figure*}[h]
    \centering
    \includegraphics[width=\textwidth]{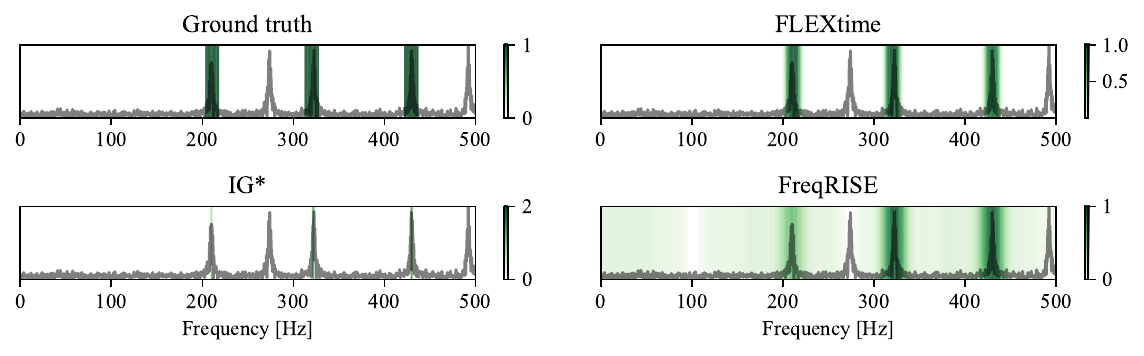}
    \caption{Examples of explanations (green heatmaps) on a synthetic dataset example.}
    \label{fig:synth_ex}
\end{figure*}

An example of explanations produced by IG, Freq\-RISE, and FLEXtime is shown in Figure \ref{fig:synth_ex}. The figure shows that IG gives very parsimonious explanations, giving less relevance to the peak at the left. On the other hand, FLEXtime marks the entire ground truth region but also exceeds beyond the bounds due to the by-design fixed bandwidth. The same is true for FreqRISE, although yielding results with a more noisy baseline. As such, FLEXtime correctly highlights relevant regions with a very clear explanation.  
\subsection{Real-life datasets}
\begin{figure*}[t]
    \centering
    \includegraphics[width=\textwidth]{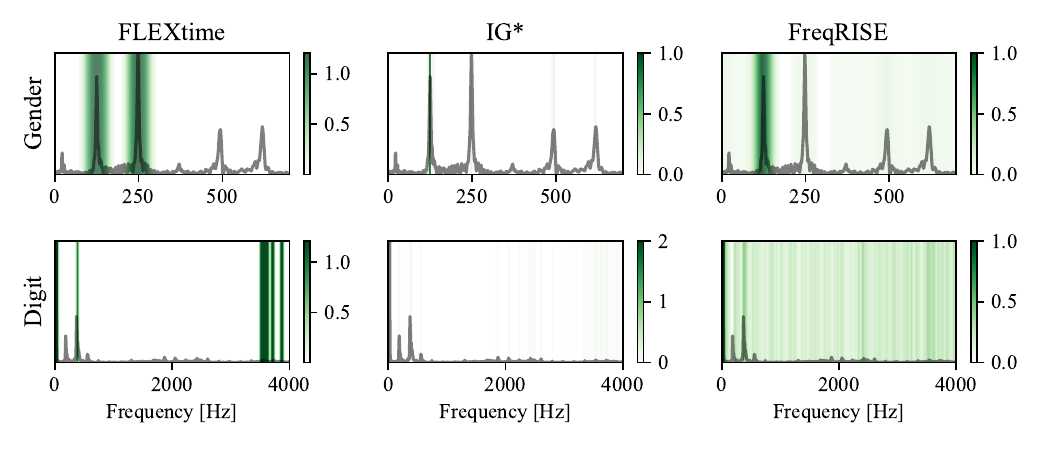}
    \caption{Explanations (green heatmaps) produced by Freq\-RISE, IG, FLEXtime on both AudioMNIST tasks. Top row is the Gender task and bottom row is the Digit task.}
    \label{fig:digit_ex}
\end{figure*} 
For the real-life datasets, the faithfulness scores are reported in Table \ref{tab:faithfulness}. The table shows that FLEXtime always gives the best or joint best faithfulness score (considering confidence intervals). Additionally, FLEXtime gives the best average rank. The competing methods are Freq\-RISE, IG, and FreqMask, however, all of them have cases where they perform significantly worse than the best method. This is the Digit task for Dynamask, SleepEDF for FreqRISE, and ECG for IG. FLEXtime gives competitive performance on all datasets. 

\begin{table*}[h]
    \small
    \centering
    \caption{Faithfulness $(\uparrow)$: Mean true class probability after inserting the 10\% most important features. We report mean and standard error of the mean in parentheses. Best result and any result with overlapping 95\% CI of the best result are bolded. Second best is underlined if CI is not overlapping with that of the best result.}
    \resizebox{\linewidth}{!}{
    \begin{tabular}{l|r|r|r|r|r|r|r}
    \toprule
         Method & \multicolumn{1}{|c|}{Gender} & \multicolumn{1}{|c|}{Digit} & \multicolumn{1}{|c|}{PAM} & \multicolumn{1}{|c|}{Epilepsy} & \multicolumn{1}{|c|}{ECG} & \multicolumn{1}{|c|}{SleepEDF} & \multicolumn{1}{|c}{Rank} \\ \hline
Random & .606(.013)& .207(.019)& .245(.007)& .819(.002)& .811(.006)& .289(.006)& 7.3 \\ \hline
Saliency* & \textbf{.943}(.024)& .414(.023)& .511(.024)& \textbf{.925}(.005)& .781(.015)& .409(.016)& 5.5\\ 
G$\times$I* & .713(.063)& .383(.011)& .516(.018)& \textbf{.880}(.027)& .793(.023)& .356(.014)& 6.3\\ 
GB* & \textbf{.947}(.025)& .442(.022)& .533(.040)& .906(.008)& .738(.018)& .388(.017)& 5.3\\ 
FreqRISE & \textbf{.970}(.013)& \underline{.858}(.010)& \textbf{.836}(.013)& \textbf{.933}(.004)& \textbf{.862}(.045)& .638(.019)& \underline{2.3}\\ 
IG* & \textbf{.970}(.014)& .604(.023)& \textbf{.814}(.025)& \textbf{.937}(.004)& .715(.067)& \textbf{.714}(.021)& 3.2\\ \hline
FreqMask & .795(.050)& .557(.025)& \textbf{.845}(.012)& .861(.013)& \textbf{.888}(.020)& .619(.022)& 4.0\\ 
FLEXtime & \textbf{.968}(.013)& \textbf{.907}(.011)& \textbf{.854}(.009)& \textbf{.928}(.003)& \textbf{.868}(.020)& \textbf{.746}(.022)& \textbf{1.8}\\ 

\bottomrule
    \end{tabular}}
    \label{tab:faithfulness}
\end{table*}

\begin{table*}[t!h]
\small
    \centering
    \caption{Complexity scores $(\downarrow)$ across all methods and datasets. We report mean and standard error of the mean in parentheses. Best result and any result with overlapping 95\% CI of the best result are bolded. Second best is underlined if CI is not overlapping with that of the best result.}
    \resizebox{\linewidth}{!}{
    \begin{tabular}{l|r|r|r|r|r|r|r}
    \toprule
         Method & \multicolumn{1}{|c|}{Gender} & \multicolumn{1}{|c|}{Digit} & \multicolumn{1}{|c|}{PAM} & \multicolumn{1}{|c|}{Epilepsy} & \multicolumn{1}{|c|}{ECG} & \multicolumn{1}{|c|}{SleepEDF}  & \multicolumn{1}{|c}{Rank}\\ \hline
Saliency* & 5.22(0.04)& 5.71(0.07)& 5.38(0.05)& 2.38(0.01)& 2.75(0.07)& 5.42(0.03)& 3.7\\ 
G$\times$I* & 5.48(0.10)& \underline{5.59}(0.04)& 5.53(0.01)& 1.59(0.24)& 2.71(0.16)& \textbf{4.88}(0.05)& 3.3\\ 
GB* & 5.40(0.12)& 6.37(0.06)& 5.46(0.04)& 2.50(0.02)& 2.83(0.01)& 5.45(0.03)& 5.5\\ 
FreqRISE & 8.00(0.02)& 8.17(0.00)& 8.51(0.00)& 4.30(0.00)& 5.10(0.00)& 7.21(0.00)& 7.0\\ 
IG* & 4.74(0.08)& 6.25(0.04)& 5.79(0.02)& 1.68(0.07)& 2.80(0.10)& \underline{5.13}(0.04)& 4.2\\ \hline
FreqMask & \textbf{1.96}(0.42)& \textbf{3.94}(0.15)& \textbf{3.58}(0.09)& \textbf{0.42}(0.05)& \textbf{0.94}(0.15)& 5.44(0.01)& \textbf{1.5}\\ 
FLEXtime & \underline{4.72}(0.44)& 6.07(0.04)& \underline{4.91}(0.11)& \underline{0.69}(0.07)& \textbf{1.28}(0.34)& 5.44(0.02)& \underline{2.7}\\ \bottomrule
    \end{tabular} }
    \label{tab:complexity}
\end{table*}

\begin{table*}[t!h]
\small
    \centering
    \caption{Smoothness scores $(\downarrow)$ across all methods and datasets. We report mean and standard error of the mean in parentheses. Best result and any result with overlapping 95\% CI of the best result are bolded. Second best is underlined if CI is not overlapping with that of the best result.}
    \resizebox{\linewidth}{!}{
    \begin{tabular}{l|r|r|r|r|r|r|r}
    \toprule
         Method & \multicolumn{1}{|c|}{Gender} & \multicolumn{1}{|c|}{Digit} & \multicolumn{1}{|c|}{PAM} & \multicolumn{1}{|c|}{Epilepsy} & \multicolumn{1}{|c|}{ECG} & \multicolumn{1}{|c|}{SleepEDF}  & \multicolumn{1}{|c}{Rank}\\ \hline
Saliency* & 30.42(0.96)& 42.53(3.23)& 13.18(0.22)& 5.05(0.04)& 6.22(0.21)& 35.21(1.12)& 5.7\\ 
G$\times$I* & 23.09(2.19)& 27.78(1.35)& 7.24(0.22)& 3.01(0.22)& 4.67(0.34)& 18.74(0.53)& 3.8\\ 
GB* & 33.12(2.74)& 84.19(5.07)& 12.69(0.25)& 5.89(0.08)& 7.34(0.30)& 37.31(1.06)& 6.3\\ 
FreqRISE & 6.70(0.61)& \textbf{10.33}(0.14)& 7.21(0.07)& 7.40(0.05)& 7.44(0.50)& \textbf{13.96}(0.54)& 4.2\\ 
IG* & 8.71(0.61)& 43.91(2.54)& 6.54(0.16)& 3.27(0.11)& 6.12(0.54)& 32.96(1.31)& 4.2\\ \hline
FreqMask & \textbf{2.49}(0.52)& \textbf{8.88}(1.12)& \textbf{0.75}(0.07)& \underline{0.78}(0.08)& \textbf{1.48}(0.28)& 42.29(1.47)& \underline{2.5}\\ 
FLEXtime & \textbf{4.11}(0.49)& \textbf{9.51}(0.18)& \textbf{0.61}(0.06)& \textbf{0.48}(0.04)& \textbf{1.00}(0.28)& \underline{16.60}(0.50)& \textbf{1.5}\\  \bottomrule
    \end{tabular} }
    \label{tab:smoothness}
\end{table*}

\begin{table*}[t!h]
\small
    \centering
    \caption{Robustness scores $(\downarrow)$ across all methods and datasets. We report mean and standard error of the mean in parentheses. Best result and any result with overlapping 95\% CI of the best result are bolded. Second best is underlined if CI is not overlapping with that of the best result.}
    \resizebox{\linewidth}{!}{
    \begin{tabular}{l|r|r|r|r|r|r|r}
    \toprule
         Method & \multicolumn{1}{|c|}{Gender} & \multicolumn{1}{|c|}{Digit} & \multicolumn{1}{|c|}{PAM} & \multicolumn{1}{|c|}{Epilepsy} & \multicolumn{1}{|c|}{ECG} & \multicolumn{1}{|c|}{SleepEDF} & \multicolumn{1}{|c}{Rank} \\ \hline
Saliency* & \textbf{14.09}(0.27)& \textbf{8.88}(1.01)& 23.61(0.34)& 17.37(0.24)& 15.35(0.41)& 18.80(0.46)& 4.2\\ 
G$\times$I* & 16.59(0.78)& \textbf{8.96}(0.50)& \textbf{16.26}(0.62)& 16.00(0.77)& \textbf{13.22}(0.76)& 15.37(0.12)& \underline{3.0}\\ 
GB* & \textbf{14.83}(0.63)& \textbf{9.56}(0.28)& 22.56(0.18)& 17.18(0.21)& 15.57(0.64)& 18.87(0.35)& 5.0\\ 
FreqRISE & 19.13(0.29)& \textbf{10.65}(0.15)& \textbf{14.88}(0.19)& 21.99(0.21)& 18.48(0.49)& \underline{14.39}(0.11)& 5.0\\ 
IG* & 16.60(0.49)& \textbf{9.43}(0.20)& 16.45(0.23)& \underline{14.65}(0.24)& \textbf{12.73}(0.67)& 15.64(0.20)& 3.5\\ \hline
FreqMask & 16.41(0.65)& 12.28(0.16)& \underline{16.18}(0.26)& 17.65(0.11)& 15.99(0.34)& 17.45(0.09)& 5.0\\ 
FLEXtime & \textbf{11.89}(1.22)& \textbf{9.34}(0.27)& 17.12(0.34)& \textbf{11.33}(0.76)& \textbf{14.50}(1.25)& \textbf{10.88}(0.41)& \textbf{2.3}\\ \bottomrule
    \end{tabular} }
    \label{tab:sensitivity}
\end{table*}

Additionally, we measure the complexities of the explanations given by the different methods. These results are reported in Table \ref{tab:complexity}. FreqMask almost always gives the least complex solution. However, in most cases FLEXtime gives the second best complexity score and overall FLEXtime yields the second best average rank. Here, we should note that while FreqMask has a lower complexity, the faithfulness is much worse than FLEXtime. As such, we must remember that neither of these metrics should be considered without the other. 


Table \ref{tab:smoothness} shows the smoothness scores. The table shows that while FreqMask and FLEXtime typically have the two best scores, FLEXtime achieves the best average rank, with FreqMask coming in second. 

Finally, the robustness scores indicate how stable the methods are against minor perturbations to the input data. Table \ref{tab:sensitivity} shows the robustness results. Again, FLEXtime has a stable, good performance. FLEXtime gets the best score on 3 datasets and is within the confidence interval on the best score on 2 other datasets. The remaining dataset, PAM, differs from the remaining datasets by having multiple channels (17 channels). This may affect the optimization procedure if some of the channels carry redundant information. 
Overall, FLEXtime again achieves the best overall rank. 

Figure \ref{fig:digit_ex} shows two examples on the AudioMNIST dataset, one for the Gender task of a male speaker, and one for the Digit task of the digit 6.
On the Gender task, all methods find the first large peak around 125Hz, adhering to expert knowledge that the male fundamental frequency typically lies in the range of 90-150Hz \cite{fitch1970a}. FLEXtime and IG additionally mark the first harmonic.
While all methods pick up on a low frequency component on the Digit task, FLEXtime also puts emphasis on the high frequency content of the signal. 
Upon closer examination, we find that when applying only the high-pass filter of the learned explanation to the signal, the model still correctly classifies the signal as the digit 6. This behavior indicates that the high frequency content is indeed relevant for the model's classification. The saliency maps of IG and FreqRISE are both more scattered, making it difficult to identify the most relevant part of the signal. This indicates that FLEXtime picks up on important components in the signal unidentified by other methods.


\subsection{Case study on sleep staging}
In this section, we perform a deeper qualitative assessment on the SleepEDF dataset. Sleep data is often described in terms of its frequency content in different sleep stages \cite{altalag2019a}. As such, we can use it to qualify whether we can discover similar patterns as described in the literature by using our explainability methods. The aim is therefore to show the benefit of using explanations in the frequency content to explain signals with known frequency-specific characteristics.
\begin{figure}[h]
    \centering
    \includegraphics[width=\textwidth]{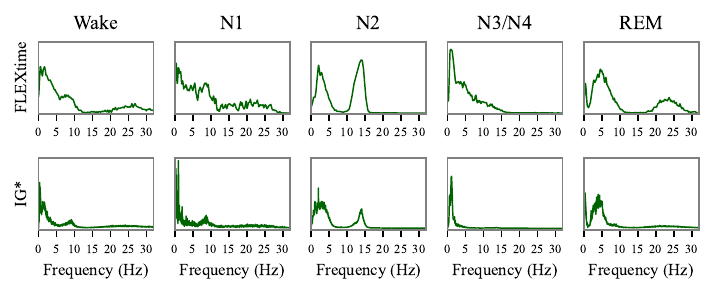}
    \caption{Average saliency maps across all correctly classified samples within each of the five classes for one of the SleepEDF test sets.}
    \label{fig:casestudy}
\end{figure}

To get an understanding of the global model behavior across an entire class, we look at one of the SleepEDF test sets. We inspect the explanations produced by FLEXtime and IG since they yield the highest faithfulness scores. We average the saliency maps produced by each method across all correctly classified samples within each class. The resulting plots are shown in Figure \ref{fig:casestudy}.

First, the figure shows that each of the classes has characteristic traces in the frequency domain, indicating that the model is indeed using specific patterns in the frequency domain to classify the data into sleep stages. IG finds that the \textit{Wake} class and the \textit{N1} sleep stage have very similar traces. For FLEXtime, they are slightly more distinguishable.
In the literature, the \textit{N1} class is characterized by mixed-frequency activity with a slight increase in the 4-7Hz band \cite{polysomnography}. This aligns with the explanations provided by both FLEXtime and IG. The \textit{Wake} class typically contains alpha activity (8-13Hz) \cite{polysomnography}, which is not apparent in IG or FLEXtime explanations. This may be due to the fact that alpha waves are typically best recorded by occipital leads \cite{altalag2019a} (EEG leads located on the back of the skull), whereas this dataset only contains the Fpz-Cz lead (EEG lead located to the front of the skull). 

The \textit{N2} sleep stage contains two clear peaks around 3 Hz and 13 Hz for both IG and FLEXtime. The \textit{N2} stage is characterized by the presence of K-complexes or sleep spindles \cite{polysomnography}. The K-complexes are mainly localized in time, whereas the sleep spindles are oscillations of 12-14Hz \cite{polysomnography} and thus align perfectly with the second peak, which is more prominent for FLEXtime. 
The \textit{N3/N4} sleep stage is also known as slow wave sleep \cite{altalag2019a}. The patterns found by both IG and FLEXtime clearly represent this with a large peak around 2Hz. 

Finally, for \textit{REM} sleep, both IG and FLEXtime have a peak at 5Hz. However, FLEXtime has a second prominent peak around 24 Hz. \textit{REM} sleep is known to contain saw-tooth waves in the 2-6 Hz range, which could explain the first peak. However, \textit{REM} sleep can also contain beta activity (16-32 Hz) \cite{merica1997a}, which could explain the second prominent peak seen in the FLEXtime explanation. 

\begin{figure}
    \centering
    \includegraphics{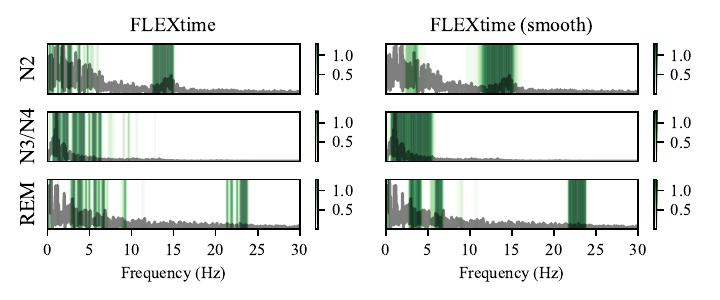}
    \caption{Examples of explanations produced by FLEXtime and a smoothness regularized version of FLEXtime on samples from the \textit{N2}, \textit{N3/N4} and \textit{REM} classes.}
    \label{fig:sleepedfexamples}
\end{figure}

Lastly, we investigate the addition of a smoothness regularizer to the FLEXtime objective. Specifically, we add the smoothness regularizer from FreqMask presented in \eqref{eq:smoothness} to the objective function in \eqref{eq:flextime}.
Figure \ref{fig:sleepedfexamples} shows examples of explanations on the \textit{N2}, \textit{N3/N4}, and the \textit{REM} class using FLEXtime with and without the smoothness regularizer. The two versions highlight similar areas, that is, for \textit{N2}, we see a prominent band around 12-14 Hz, for \textit{N3/N4} the importance is concentrated on low frequency content (<5Hz), and for \textit{REM} there is a highlighted section around 23 Hz. All of this is consistent with the characteristics mentioned previously. However, the smoothness regularized version of FLEXtime removes some of the noise, making the explanations more easily comprehendible. When evaluating the smoothness regularized version of FLEXtime across the entire test set and across the five seeds, we achieve better complexity, $5.38(0.01)$, and smoothness, $7.04(0.11)$. However, faithfulness drops to $0.724(0.021)$.

\section{Discussion and conclusion}
We proposed to reimagine time series as explainability over interpretable parts. In particular, we presented FLEXtime, an approach to time series explainability that utilizes the powerful notion of filterbanks to learn relevant explanations in the frequency domain. 

FLEXtime outperforms state-of-the-art baselines on faithfulness, smoothness and robustness on a range real-life datasets. The datasets cover a variety of domains and comprise time series of variable signal length and number of variables. FLEXtime also has competitive performance on localization. 
Simultaneously, FLEXtime outperforms the simpler baseline FreqMask also developed for this study. This is likely because FLEXtime reduces the dimensionality of the problem to make for easier optimization by leveraging the fact that frequency information is often localized in bands. 

Additionally, due to the filtering approach, FLEXtime is able to handle streaming time series more smoothly, compared to the remaining methods which would have to rely on windowing to transform the data into the frequency domain.

\textbf{Faithfulness vs. complexity}
FLEXtime achieves the best rank on the faithfulness score, but only second best on the complexity score, where it is outperformed by FreqMask. It is, however, important to note that the complexity can only be interpreted while simultaneously considering the faithfulness scores since a sparse solution is irrelevant if it does not highlight the correct regions. E.g., while FLEXtime has higher complexity score than FreqMask on the Digit task, it also achieves the best faithfulness score. This indicates that a more complex solution is needed to adequately describe this dataset.

\textbf{Towards different filter bank designs}
In this work, we used FIR filterbanks with parameters tuned through cross-validation to model our explanations. This means that not all filterbanks are able to perfectly reconstruct the signals. This will rarely be an issue since we are looking for sparse masks that highlight relevant regions of interest. However, we believe that future work should focus on investigating the effect of perfect reconstruction filterbanks on the results \cite{vaidyanathan1987a}. Additionally, using tailored filterbanks for the different domains will allow domain experts to utilize prior knowledge to extract meaningful explanations from the data \cite{chandra2020a,mittal2021a}. Finally, considering alternative ways to split the signals into interpretable parts may shed light on new avenues for time series explainability. 

\textbf{Metrics and qualitative assessment}
Quantifying the quality of the explanations is difficult. This, for instance, becomes apparent when considering FLEXtime with and without smoothness regularization. On the SleepEDF dataset, the smoothness regularized version gives an explanation that is easier to comprehend, but with a lower faithfulness. This may be due to the computation of faithfulness, which assigns pointwise importance in the frequency domain. This approach will likely favor sharper and more precise explanations at the cost of a more noisy looking importance map. 
Here, we have optimized for faithfulness, but we find that optimizing for other metrics yields different results.
We therefore believe future work should investigate how to properly quantify and balance different qualities when building new explainability methods for time series. 

\textbf{Case-study on sleep data} Using FLEXtime we were able to identify known markers for different sleep stages. This shows the benefit of moving to the frequency domain, where we can use known identifiers of class-specific characteristics to explain the models. Confirming existing knowledge using FLEXtime is a valuable tool for model debugging and validation. However, a more exciting avenue for explainability in general is to shed new light in understanding underlying mechanisms in the data. We hope that the case study can increase trust in FLEXtime and be a tool for domain experts to formulate new hypotheses for gaining a deeper understanding of disease patterns using data such as EEG.

\textbf{Limitations} Although FLEXtime has clear competitive advantages across a range of metrics, it still comes with potential for future work. One current limitation of the method is that it requires tuning or choice of hyperparameters for the filterbank. Especially, if no expert knowledge is available. Automatic filterbank design could be a fruitful future direction for development. Additionally, as of now, the filterbank design of FLEXtime offers only insights into the frequency domain, and no explanation of time relevant features. Future work should focus on exploring the trade-off between the two and potentially incorporating insights from both domains via new filterbank designs.

\subsection{Conclusion}
Our new method, FLEXtime, bridges an important gap in time series explainability, where multiple works have successfully applied learnable masks in the time domain, but none in the frequency domain. Additionally, we highlight that naive masking over the DFT, in effect, means that we would mask out single frequencies, and this goes against established signal processing theory. FLEXtime instead learns the mask over a sufficiently expressive filterbank. This leads to a much more computationally stable procedure with a clear learning objective while having the flexibility to tailor the filterbank to the application at hand. This point is furthered by the superior performance of FLEXtime over FreqMask.
Combined with the competitive performance of FLEXtime, outperforming all established state-of-the-art baselines across a range of metrics, we hope that this work will inspire future research in this direction. 

\section{Supplementary material}
\subsection{Filtering by zeroing out frequency components} \label{sec:artifacts}
\begin{figure}[h]
    \centering
    \includegraphics[width=0.7\textwidth]{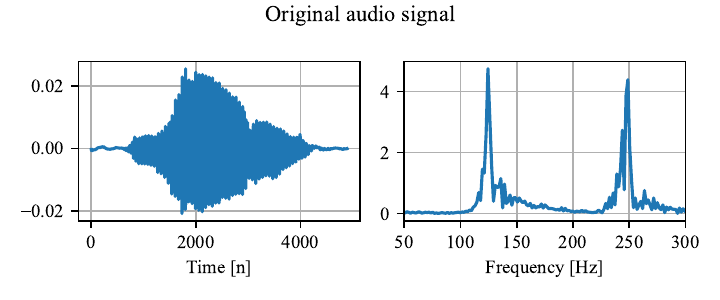}
    \caption{Digit \textit{one} spoken by a male speaker in the time and frequency domain.}
    \label{fig:supp_origdata}
\end{figure}

Filters by zeroing components (DFT filtering) will introduce artifacts, as an ideal filter is not realizable \cite{oppenheim2010a} (Chapter 4).
To demonstrate this behavior, we consider an example from the AudioMNIST dataset \cite{audiomnist}, where the digit \textit{one} is spoken by a male speaker. Figure \ref{fig:supp_origdata} shows the data example in the time domain and part of the frequency domain. The signal has a significant harmonic in the band from $124-130$ Hz.

\begin{figure}[h]
    \centering
    \includegraphics[width=0.7\textwidth]{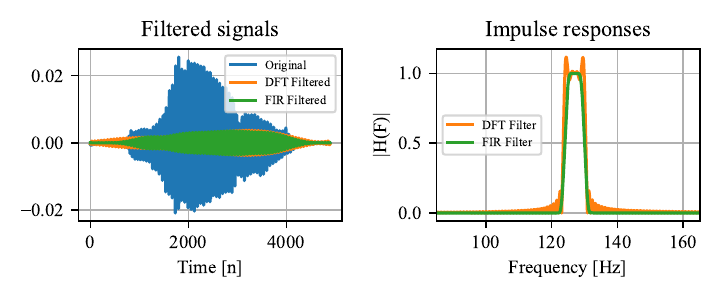}
    \caption{Left: Example from AudioMNIST, where the signal is DFT filtered and FIR filtered. Right: Impulse response produced by DFT filter and FIR filter.}
    \label{fig:supp_filter_example}
\end{figure}

We now bandpass the signal in the range $124-130$ Hz using two different approaches; DFT filtering (i.e. setting the frequency components outside of the range to 0) and filtering using an FIR filter. The two approaches, along with the original signal, are shown in Figure \ref{fig:supp_filter_example} (left).
Here, we see that the DFT filtering is introducing artifacts in the filtered signal, especially from sample 0 to 2800. The FIR filtering does not introduce any signal artifacts and instead only filters the signal in the desired frequency range.

This phenomenon can also be understood in the frequency domain. Zeroing frequency components will approximate an ideal filter, but by doing so, oscillations are created close to the transient frequencies. This is illustrated in Figure \ref{fig:supp_filter_example} (right). 
Here, the DFT filter and the FIR filter are of the same length, and we clearly see that the DFT filter introduces a greater amount of artifacts compared to the FIR filter. This behavior is also referred to as the Gibbs phenomenon \cite{oppenheim2010a}.

The phenomenon occurs because of the difficulty of approximating the jump discontinuities in the ideal bandpass filter using a Fourier series, which is essentially, what the DFT filter does.

\begin{credits}
\subsubsection{\ackname} This work was supported by the Pioneer Centre for AI, DNRF grant number P1, as well as the Research Council of Norway via Visual Intelligence grant no. 309439 and grant no. 303514.

\subsubsection{\discintname}
The authors have no competing interests to declare that are relevant to the content of this article. 
\end{credits}
%
%
%
\bibliographystyle{splncs04}
\bibliography{refs}

\begin{thebibliography}{10}
\providecommand{\url}[1]{\texttt{#1}}
\providecommand{\urlprefix}{URL }
\providecommand{\doi}[1]{https://doi.org/#1}

\bibitem{achtibat2023a}
Achtibat, R., Dreyer, M., Eisenbraun, I., Bosse, S., Wiegand, T., Samek, W., Lapuschkin, S.: From attribution maps to human-understandable explanations through concept relevance propagation. Nature Machine Intelligence  \textbf{5}(9) (2023)

\bibitem{agarwal2022rethinking}
Agarwal, C., Johnson, N., Pawelczyk, M., Krishna, S., Saxena, E., Zitnik, M., Lakkaraju, H.: Rethinking stability for attribution-based explanations. In: ICLR 2022 Workshop on PAIR{\textasciicircum}2Struct (2022)

\bibitem{altalag2019a}
Altalag, A., Road, J., Wilcox, P., Aboulhosn, K.: Pulmonary Function Tests in Clinical Practice, chap. Diagnostic tests for sleep disorders. Springer International Publishing, 2nd edn. (2019)

\bibitem{faithfulness2}
Alvarez-Melis, D., Jaakkola, T.S.: Towards robust interpretability with self-explaining neural networks. In: Proceedings of the 32nd International Conference on Neural Information Processing Systems. NIPS'18 (2018)

\bibitem{amann2020a}
Amann, J., Blasimme, A., Vayena, E., Frey, D., Madai, V.I.: Explainability for artificial intelligence in healthcare: a multidisciplinary perspective. Bmc Medical Informatics and Decision Making  \textbf{20}(1), ~310 (2020)

\bibitem{gxi}
Ancona, M., Ceolini, E., Öztireli, C., Gross, M.: Towards better understanding of gradient-based attribution methods for deep neural networks. In: International Conference on Learning Representations (2018)

\bibitem{epilepsy}
Andrzejak, R., Lehnertz, K., Mormann, F., Rieke, C., David, P., Elger, C.: Indications of nonlinear deterministic and finite-dimensional structures in time series of brain electrical activity: Dependence on recording region and brain state. Physical review. E, Statistical, nonlinear, and soft matter physics  \textbf{64} (2002)

\bibitem{loc2}
Arias-Duart, A., Parés, F., Garcia-Gasulla, D., Giménez-Ábalos, V.: Focus! rating xai methods and finding biases. In: IEEE International Conference on Fuzzy Systems (2022)

\bibitem{loc1}
Arras, L., Osman, A., Samek, W.: {CLEVR-XAI: A benchmark dataset for the ground truth evaluation of neural network explanations}. Information Fusion  \textbf{81} (2022)

\bibitem{lrp}
Bach, S., Binder, A., Montavon, G., Klauschen, F., Müller, K.R., Samek, W.: On pixel-wise explanations for non-linear classifier decisions by layer-wise relevance propagation. PLOS ONE  \textbf{10}(7),  1--46 (07 2015)

\bibitem{audiomnist}
Becker, S., Vielhaben, J., Ackermann, M., Müller, K.R., Lapuschkin, S., Samek, W.: {AudioMNIST}: Exploring explainable artificial intelligence for audio analysis on a simple benchmark. Journal of the Franklin Institute  \textbf{361}(1) (2024)

\bibitem{beger2024a}
Beger, J.: The crucial role of explainability in healthcare {AI}. European Journal of Radiology  \textbf{176},  111507 (2024)

\bibitem{bhatt2020a}
Bhatt, U., Weller, A., Moura, J.M.: Evaluating and aggregating feature-based model explanations. IJCAI International Joint Conference on Artificial Intelligence  (2020)

\bibitem{impossibility}
Bilodeau, B., Jaques, N., Koh, P.W., Kim, B.: Impossibility theorems for feature attribution. Proceedings of the National Academy of Sciences  \textbf{121}(2) (2024)

\bibitem{freqrise}
Br{\"u}sch, T., Wickstr{\o}m, K.K., Schmidt, M.N., Alstr{\o}m, T.S., Jenssen, R.: Freq{RISE}: Explaining time series using frequency masking. In: Northern Lights Deep Learning Conference (2025)

\bibitem{complexity2}
Chalasani, P., Chen, J., Chowdhury, A.R., Wu, X., Jha, S.: Concise explanations of neural networks using adversarial training. In: Proceedings of the 37th International Conference on Machine Learning. PMLR (2020)

\bibitem{chandra2020a}
Chandra, S., Sharma, A., Singh, G.K.: Computationally efficient cosine modulated filter bank design for {ECG} signal compression. Irbm  \textbf{41}(1) (2020)

\bibitem{cover1991a}
Cover, T.M., Thomas, J.A.: Elements of Information Theory, 2nd edition, chap.~10. Wiley (2005)

\bibitem{dynamask}
Crabb{\'e}, J., Van Der~Schaar, M.: Explaining time series predictions with dynamic masks. In: Meila, M., Zhang, T. (eds.) Proceedings of the 38th International Conference on Machine Learning. PMLR (2021)

\bibitem{crabbé2024time}
Crabbé, J., Huynh, N., Stanczuk, J., van~der Schaar, M.: Time series diffusion in the frequency domain (2024)

\bibitem{crochiere1983multirate}
Crochiere, R.E., Rabiner, L.R.: Multirate digital signal processing, chap.~7. Prentice-hall Englewood Cliffs, NJ (1983)

\bibitem{di2023a}
Di~Martino, F., Delmastro, F.: Explainable ai for clinical and remote health applications: a survey on tabular and time series data. Artificial Intelligence Review  \textbf{56}(6) (2023)

\bibitem{sparsedenoising}
Elad, M., Aharon, M.: Image denoising via sparse and redundant representations over learned dictionaries. IEEE Transactions on Image Processing  \textbf{15}(12) (2006)

\bibitem{extrmask}
Enguehard, J.: Learning perturbations to explain time series predictions. In: Proceedings of the 40th International Conference on Machine Learning. Proceedings of Machine Learning Research, vol.~202. PMLR (2023)

\bibitem{fitch1970a}
Fitch, J., Holbrook, A.: Modal vocal fundamental frequency of young adults. Archives of Otolaryngology  \textbf{92}(4) (1970)

\bibitem{fong2019a}
Fong, R., Patrick, M., Vedaldi, A.: Understanding deep networks via extremal perturbations and smooth masks. IEEE/CVF International Conference on Computer Vision  (2019)

\bibitem{PhysioNet}
Goldberger, A.L., Amaral, L.A., Glass, L., Hausdorff, J.M., Ivanov, P.C., Mark, R.G., Mietus, J.E., Moody, G.B., Peng, C.K., Stanley, H.E.: Physiobank, physiotoolkit, and physionet: components of a new research resource for complex physiologic signals. Circulation  \textbf{101}(23) (2000)

\bibitem{hedstrom2023quantus}
Hedstr{\"{o}}m, A., Weber, L., Krakowczyk, D., Bareeva, D., Motzkus, F., Samek, W., Lapuschkin, S., H{\"{o}}hne, M.M.M.: Quantus: An explainable {AI} toolkit for responsible evaluation of neural network explanations and beyond. Journal of Machine Learning Research  \textbf{24}(34) (2023)

\bibitem{hedstrom2023metaquantus}
Hedström, A., Bommer, P., Wickstrøm, K.K., Samek, W., Lapuschkin, S., Höhne, M.M.C.: The meta-evaluation problem in explainable {AI}: Identifying reliable estimators with {MetaQuantus}. Transactions on Machine Learning Research  (2023)

\bibitem{polysomnography}
Jafari, B., Mohsenin, V.: Polysomnography. Clinics in Chest Medicine  \textbf{31}(2) (2010)

\bibitem{sleepedf}
Kemp, B., Zwinderman, A.H., Tuk, B., Kamphuisen, H.A., Oberyé, J.J.: Analysis of a sleep-dependent neuronal feedback loop: The slow-wave microcontinuity of the {EEG}. IEEE Transactions on Biomedical Engineering  \textbf{47}(9) (2000)

\bibitem{kim2010sparse}
Kim, T., Shakhnarovich, G., Urtasun, R.: Sparse coding for learning interpretable spatio-temporal primitives. Advances in neural information processing systems  \textbf{23} (2010)

\bibitem{Kolek2022}
Kolek, S., Nguyen, D.A., Levie, R., Bruna, J., Kutyniok, G.: A Rate-Distortion Framework for Explaining Black-Box Model Decisions. Springer International Publishing (2022)

\bibitem{frequency-aware-masked-autoencoders}
Liu, R., Zippi, E., Ansari, H.P., Sandino, C., Nie, J., Goh, H., Azemi, E., Moin, A.: Frequency-aware masked autoencoders for multimodal pretraining on biosignals. In: ICLR Workshop (2024)

\bibitem{timexplusplus}
Liu, Z., Wang, T., Shi, J., Zheng, X., Chen, Z., Song, L., Dong, W., Obeysekera, J., Shirani, F., Luo, D.: Timex++: Learning time-series explanations with information bottleneck. In: Proceedings of the 41st International Conference on Machine Learning (2024)

\bibitem{contralsp}
Liu, Z., Zhang, Y., Wang, T., Wang, Z., Luo, D., Du, M., Wu, M., Wang, Y., Chen, C., Fan, L., Wen, Q.: Explaining time series via contrastive and locally sparse perturbations. In: The International Conference on Learning Representations (2024)

\bibitem{shap}
Lundberg, S.M., Lee, S.I.: A unified approach to interpreting model predictions. In: Proceedings of the 31st International Conference on Neural Information Processing Systems. NIPS'17, Curran Associates Inc., Red Hook, NY, USA (2017)

\bibitem{mercier2022a}
Mercier, D., Dengel, A., Ahmed, S.: {TimeREISE}: Time series randomized evolving input sample explanation. Sensors  \textbf{22}(11), ~4084 (2022)

\bibitem{merica1997a}
Merica, H., Blois, R.: Relationship between the time courses of power in the frequency bands of human sleep eeg. Neurophysiologie Clinique  \textbf{27}(2) (1997)

\bibitem{mittal2021a}
Mittal, R., Prince, A.A., Nalband, S., Robert, F., Fredo, A.R.J.: Modified-mamemi filter bank for efficient extraction of brainwaves from electroencephalograms. Biomedical Signal Processing and Control  \textbf{69} (2021)

\bibitem{mitecg}
Moody, G., Mark, R.: The impact of the {MIT-BIH} arrhythmia database. IEEE Engineering in Medicine and Biology Magazine  \textbf{20}(3) (2001)

\bibitem{10095297}
Morante, M., Østergaard, J., Theodoridis, S.: Interpretable nonnegative incoherent deep dictionary learning for {FMRI} data analysis. In: 2023 IEEE International Conference on Acoustics, Speech and Signal Processing (ICASSP) (2023)

\bibitem{windowshap}
Nayebi, A., Tipirneni, S., Reddy, C.K., Foreman, B., Subbian, V.: {WindowSHAP}: An efficient framework for explaining time-series classifiers based on shapley values. Journal of Biomedical Informatics  (2023)

\bibitem{oppenheim2010a}
Oppenheim, A.V., Schafer, R.W.: Discrete-time signal processing. Pearson Education, 3rd edn. (2010)

\bibitem{rise}
Petsiuk, V., Das, A., Saenko, K.: {RISE}: Randomized input sampling for explanation of black-box models. British Machine Vision Conference 2018, BMVC 2018  (2019)

\bibitem{proakis1992a}
Proakis, J., Manolakis, D.: Digital signal processing, chap.~11. Macmillan, 4th edn. (2007)

\bibitem{timex}
Queen, O., Hartvigsen, T., Koker, T., Huan, H., Tsiligkaridis, T., Zitnik, M.: Encoding time-series explanations through self-supervised model behavior consistency. In: Proceedings of Neural Information Processing Systems, NeurIPS (2023)

\bibitem{pam}
Reiss, A., Stricker, D.: Introducing a new benchmarked dataset for activity monitoring. In: 2012 16th International Symposium on Wearable Computers (2012)

\bibitem{lime}
Ribeiro, M.T., Singh, S., Guestrin, C.: "{Why} should i trust you?": {Explaining} the predictions of any classifier. In: Proceedings of the 22nd ACM SIGKDD International Conference on Knowledge Discovery and Data Mining (2016)

\bibitem{rojat2021a}
Rojat, T., Puget, R., Filliat, D., Del~Ser, J., Gelin, R., Díaz-Rodríguez, N.: Explainable artificial intelligence {(XAI)} on timeseries data: A survey  (2021)

\bibitem{schlegel2019towards}
Schlegel, U., Arnout, H., El-Assady, M., Oelke, D., Keim, D.A.: Towards a rigorous evaluation of xai methods on time series. In: 2019 IEEE/CVF International Conference on Computer Vision Workshop (ICCVW). IEEE (2019)

\bibitem{schroeder2023a}
Schröder, M., Zamanian, A., Ahmidi, N.: Post-hoc saliency methods fail to capture latent feature importance in time series data. Lecture Notes in Computer Science  \textbf{13932} (2023)

\bibitem{saliency}
Simonyan, K., Vedaldi, A., Zisserman, A.: Deep inside convolutional networks: Visualising image classification models and saliency maps (2014)

\bibitem{limesegment}
Sivill, T., Flach, P.: {LIMESegment}: Meaningful, realistic time series explanations. In: Camps-Valls, G., Ruiz, F.J.R., Valera, I. (eds.) Proceedings of The 25th International Conference on Artificial Intelligence and Statistics. Proceedings of Machine Learning Research, PMLR (2022)

\bibitem{guided-backprop}
Springenberg, J.T., Dosovitskiy, A., Brox, T., Riedmiller, M.: Striving for simplicity: The all convolutional net (2015)

\bibitem{integratedgradients}
Sundararajan, M., Taly, A., Yan, Q.: Axiomatic attribution for deep networks. In: Proceedings of the 34th International Conference on Machine Learning. Proceedings of Machine Learning Research, vol.~70. PMLR (2017)

\bibitem{Tepper_Garc_a_2006}
Tepper~García, T.: Voigt profile fitting to quasar absorption lines: an analytic approximation to the voigt-hjerting function: A new method to compute voigt profiles. Monthly Notices of the Royal Astronomical Society  \textbf{369}(4) (2006)

\bibitem{theodoridis2020a}
Theodoridis, S.: Machine Learning: A Bayesian and Optimization Perspective, Second Edition, chap.~9. Elsevier (2020)

\bibitem{tolooshams2022stable}
Tolooshams, B., Ba, D.E.: Stable and interpretable unrolled dictionary learning. Transactions on Machine Learning Research  (2022)

\bibitem{vaidyanathan1987a}
Vaidyanathan, P.P.: Quadrature mirror filter banks, m-band extensions and perfect-reconstruction techniques. Ieee Assp Magazine  \textbf{4}(3) (1987)

\bibitem{vilrp}
Vielhaben, J., Lapuschkin, S., Montavon, G., Samek, W.: Explainable {AI} for time series via {Virtual Inspection Layers}. Pattern Recognition  \textbf{150} (2024)

\bibitem{weber2024a}
Weber, P., Carl, K.V., Hinz, O.: Applications of explainable artificial intelligence in finance—a systematic review of finance, information systems, and computer science literature. Management Review Quarterly  \textbf{74}(2) (2024)

\bibitem{relax}
Wickstrøm, K.K., Trosten, D.J., Løkse, S., Boubekki, A., Mikalsen, K.O., Kampffmeyer, M.C., Jenssen, R.: {RELAX}: Representation learning explainability. International Journal of Computer Vision  \textbf{131}(6) (2023)

\bibitem{7457891}
Zhang, X., Lin, W., Ma, S., Wang, S., Gao, W.: Rate-distortion based sparse coding for image set compression. In: Visual Communications and Image Processing (2015)

\end{thebibliography}

\end{document}